\documentclass[11pt,a4paper]{article}
\usepackage[hyperref]{acl2021}
\usepackage{times}
\usepackage{latexsym}

\usepackage{microtype}
\usepackage[utf8]{inputenc}
\usepackage{graphicx}
\usepackage{multirow}
\usepackage{booktabs}
\usepackage{amsmath}
\usepackage{comment}
\usepackage{amssymb}
\usepackage{url}
\usepackage{rotating}

\aclfinalcopy 

\newcommand{\MC}{\multicolumn}

\newcommand{\xsim}{\texttt{xsim}}

\title{(Self-) Supervised Language Distillation \\ of Multilingual Sentence Representations}
\title{Joint Teacher-Student and Masked LM Training \\ for Very-Low Resource Languages}
\title{Bitext Mining Using Distilled Sentence Representations\\ for  Low-Resource Languages}
\author{
  Kevin Heffernan \and Onur Çelebi \and  Holger Schwenk \\
  \texttt{kevinheffernan@fb.com}
  \and
  \texttt{celebio@fb.com}
  \and
  \texttt{schwenk@fb.com} \\
  Meta AI Research
  }

\date{}

\newcommand{\nbLangLaser}{93}
\newcommand{\nbLangLabse}{109}
\newcommand{\nbLangLabseAfric}{13}

\newcommand{\nbLangAfric}{50}

\newcommand{\nbLangWMT}{24}
\newcommand{\nbLangFlores}{44}
\newcommand{\nbLangAddtl}{6}
\newcommand{\nbLangLASERimproved}{12}
\newcommand{\monoENG}{21.5 billion}

\begin{document}
\maketitle
\begin{abstract}

Scaling multilingual representation learning beyond the hundred most frequent languages is challenging, in particular to cover the long tail of low-resource languages. A promising approach has been to train one-for-all multilingual models capable of cross-lingual transfer, but these models often suffer from insufficient capacity and interference between unrelated languages. Instead, we move away from this approach and focus on training multiple language (family) specific representations, but most prominently enable all languages to still be encoded in the same representational space. To achieve this, we focus on teacher-student training, allowing all encoders to be mutually compatible for bitext mining, and enabling fast learning of new languages.
We introduce a new teacher-student training scheme which combines supervised and self-supervised training, allowing encoders to take advantage of monolingual training data, which is valuable in the low-resource setting.

Our approach significantly outperforms the original LASER encoder. We study very low-resource languages and handle \nbLangAfric{} African languages, many of which are not covered by any other model. For these languages, we train sentence encoders, mine bitexts, and validate the bitexts by training NMT systems. 
\end{abstract}

\section{Introduction}

There is increasing interest in multilingual sentence representations since they promise an appealing approach to extend many NLP tasks to a large number of languages, without the need to separately train a language-specific model.
Most of the current works on multilingual sentence representations have focused on training one model which handles all languages of interest, e.g. \citep{Artetxe:2019:tacl_massive_ml,google:2020:labse,Reimers:2020:emnlp_destil,samantar:2022:tacl}. The main motivation is of course that languages with limited resources will benefit from the fact that the same model has learned other (similar) languages.
Zero-shot performance is of particular interest: the model generalizes well to a new language although it has never seen training data in that language.
Training massively multilingual models faces several problems with increasing number of languages: how to make sure that all languages are learned, how to to account for the large imbalance of available training data when determining the (joint) vocabulary and during the training process itself, or the high computational complexity.

\begin{figure*}[t!]
    \centering
    \includegraphics[width=0.8\textwidth]{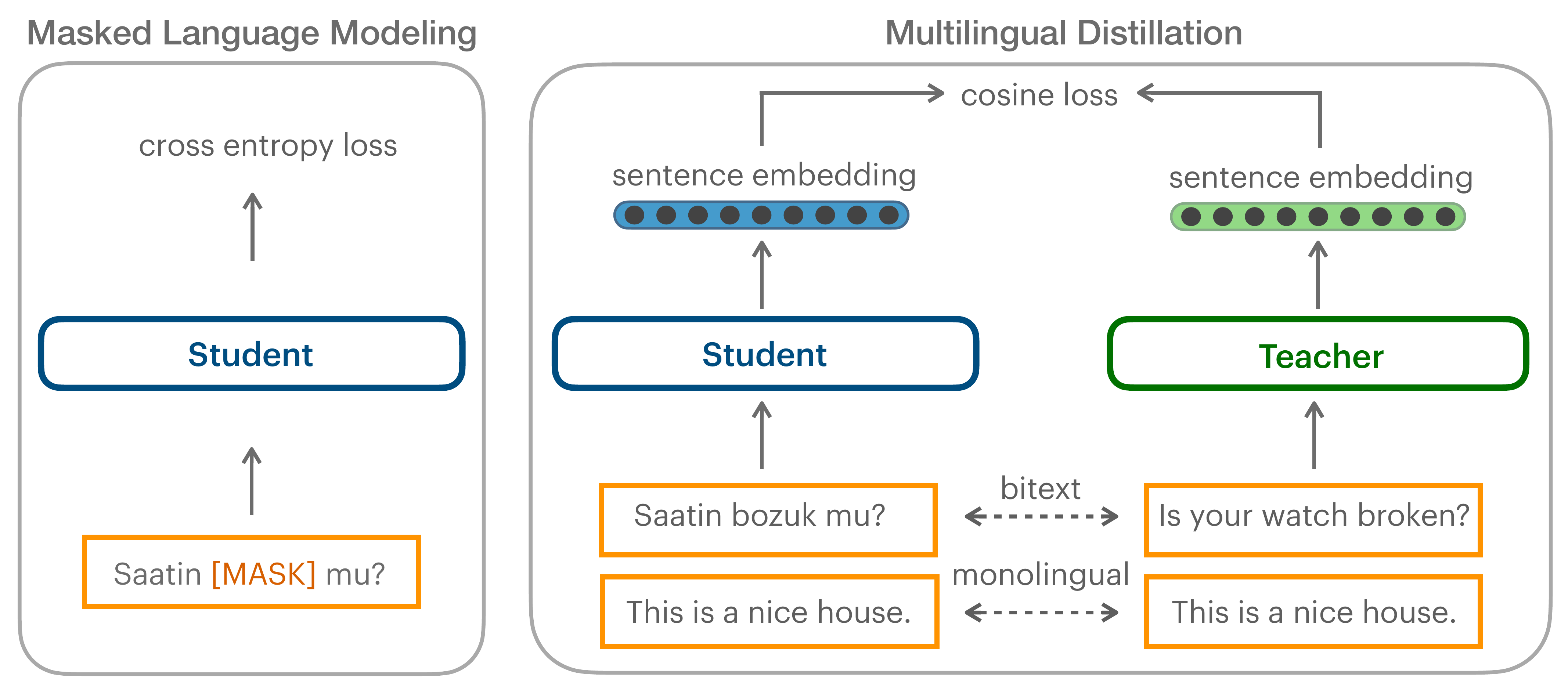}
    \caption{Architecture of our teacher-student approach.}
    \label{fig:archi}
\end{figure*}

Instead of training a massively multilingual sentence encoder from scratch, \citet{Reimers:2020:emnlp_destil} proposed a teacher-student approach to extend an existing (monolingual) sentence embedding space to new languages. We build on this generic idea and propose multiple improvements which significantly improve performance, namely different teacher and student architectures, several supervised and unsupervised training criterion, and language-specific encoders. We also investigate challenges when handling low-resources languages, showcased by training models for \nbLangAfric{} African languages. To the best of our knowledge, many of these languages are not handled by any other sentence encoder or pretrained model.
For these languages, we train sentences encoders, mine bitexts against \monoENG{} English sentences, and train
SMT models to translate into Enlish.

Multilingual sentence embeddings have many applications which is reflected by several approaches to evaluate them. Many task-specific evaluation metrics are summarized in the XTREME bench mark \citep{google:2020:icml_xtreme, google:2021:emnlp_xtremer}.
In this work, we focus on the use of multilingual sentence embeddings for similarity-based bitext mining, as proposed by \citet{Artetxe:2019:mine_acl}, and on using these mined bitexts to improve NMT. Consequently, our primary metric is NMT performance.
However, mining and NMT training is computationally expensive and it is intractable to systematically perform this evaluation for many different sentence encoder variants. As an evaluation proxy, we use multilingual similarity search error rate. In contrast to previous work which used the Tatoeba test set, e.g. \citep{Artetxe:2019:tacl_massive_ml,hu:2020:arxiv_xtreme,Reimers:2020:emnlp_destil}, we switch to the FLORES evaluation benchmark, which contains high-quality human translated texts from Wikipedia \citep{flores101:2021} and covers many low-resource languages.

The contributions of this work can be summarized as follows:
\begin{itemize}
    \item we move away from the popular \textit{one-for-all approach} and train multiple, mutually compatible language (family) specific encoders;
    \item we explore several variants and improvements of teacher-student training for multilingual sentence representations, and propose a new approach which combines supervised teacher-student with self-supervised MLM training to better handle very low-resource languages;
    \item the new model substantially improves \nbLangLASERimproved{} languages which were badly handled by the original LASER encoder;
    \item we train sentence encoders for \nbLangAfric{} African languages, mine bitexts, and train NMT systems. To the best of our knowledge, many of these languages are not handled by any other NMT system;
\end{itemize}

\noindent This paper is structured as follows. In the next section, we first summarize related work. We then describe our approach in \autoref{sec:approach} and discuss differences to existing works.
The experimental evaluation is divided into two sections: we first analyze different training techniques and evaluate the similarity search error rate (\autoref{sec:eval_xsim}).
We then switch to the challenging task of training sentence encoders and perform mining for many African languages (\autoref{sec:eval_afric}). The paper concludes with a discussion.

\section{Related work}

\paragraph{Multilingual sentence representation}
There is a large body of research on learning multilingual representations. Examples of such approaches are multilingual BERT (m-BERT) which covers 104 languages \citep{devlin-etal-2019-bert}, XLM \citep{conneau:2019:nips_xlm}, and XLM-R which was trained on 100 languages using crawled web data \citep{conneau:2020:acl_xlm_scale}. However, as these approaches do not take into account a sentence-level objective during training, they can result in poor performance when applied to tasks which use sentence representations such as bitext retrieval \citep{hu:2020:arxiv_xtreme}. In order to address this, methods such as SentenceBERT (SBERT) make use of a Siamese network to better model sentence representations \citep{Reimers:2019:emnlp_sent_bert}.
LaBSE \citep{google:2020:labse} uses a dual-encoder approach with a transformer-based architecture and additive margin softmax loss \citep{Yang:2019:ijcai_add_marg_softmax}. It covers 109 languages, and is pre-trained using a masked language modelling (MLM) and translation language modelling objective \citep{conneau:2020:acl_xlm_scale}.
LabSE was used to mine bitexts in eleven Indian languages \cite{samantar:2022:tacl}.
Another popular multilingual sentence embedding model is LASER \citep{Artetxe:2019:tacl_massive_ml}. It is based on a LSTM encoder/decoder architecture with a fixed-size embedding layer and no pre-training. LASER covers 93 languages.

When learning a multilingual embedding space, a limitation of many existing approaches is that they require training a new model from scratch each time a language is to be added. However, there have been various methods proposed to address this. \citet{Wang:2020:arxiv_xmBERT} provide one such technique which extends m-BERT to low-resource languages by increasing the size of the existing vocabulary, and then continuing self-supervised training using monolingual data for a low-resource language. Another example by \citet{Reimers:2020:emnlp_destil} uses multilingual distillation. In this supervised teacher-student approach, the teacher is a monolingual model pre-trained on English (SBERT), and the student is a pre-trained multilingual model (XLM-R). Using bitexts, the student then extends the  embedding space to the desired language(s) by applying regression loss between the English sentence representation of the teacher, and the target language sentence representation of the student. 

\paragraph{Scaling multilinguality}
Several recent works have addressed the challenges faced when scaling multilingual models to a hundred languages and beyond, namely massively multilingual NMT systems \citep{fan:2020:jalt_m2m,google:2019:arxiv_xnmt_in_the_wild}. A recent study explored the extension to more than a thousand languages \citep{Siddhant:2022:arxiv_next1000xnmt,google:2022:arxiv_nmt1500}.
Training NMT models for a large number of languages faces many challenges and a large variety of architectures have been explored \citep{deltalm:2021:arxiv,deepnet:2022:arxiv,escolano-etal-2021-multilingual}.
To the best of our knowledge, similar modelling techniques were not yet considered to train (massively) multilingual sentence encoders.

\paragraph{Resources for African languages}
The Masakhane project\footnote{\url{https://www.masakhane.io/}} aims at providing resources to both strengthen and spur NLP research in African languages.
A workshop focused on the evaluation of African languages will be held at EMNLP'22.\footnote{\url{https://www.statmt.org/wmt22/large-scale-multilingual-translation-task.html}}
In the framework of the data track, several parallel corpora were made available.
In general, the number of languages covered is in the twenties, well below the \nbLangFlores{} we evaluate in this work.

\section{Architecture}
\label{sec:approach}

The overall architecture of our approach is summarized in \autoref{fig:archi}. The teacher is an improved LASER encoder.\footnote{\url{https://github.com/facebookresearch/LASER}} Compared to the original training procedure described in \citet{Artetxe:2019:tacl_massive_ml}, we use SPM instead of BPE preprocessing, up-sampling of low-resource languages, and a new implementation in \texttt{fairseq}. This training code will be freely available in the fairseq github repository.
All the other parameters are unchanged, namely a 5-layer BiLSTM encoder, the 1024 dimensional sentence embeddings are obtained by max-pooling over the last layer, and training is performed for 93 languages with public resources obtained from OPUS. The reader is referred to \citet{Artetxe:2019:tacl_massive_ml} for details on the original LASER training procedure.
We use this new multilingual sentence encoder as the teacher in all our experiments and in this work refer to our teacher as LASER2, and student models as LASER3.

Training of the students follows the general idea of a teacher-student approach as initially proposed by \citet{Reimers:2020:emnlp_destil}, but with several important differences.
We want to scale encoder training and bitext mining well beyond the roughly 100 languages handled by current multilingual encoders. This may include languages which are not covered by existing pretrained models, and retraining them would be computationally very expensive. Also, those languages may be written in a new script which is not covered.
Therefore, we made the following design choices:
\begin{itemize}
    \item We do not initialize the student with some pretrained model, e.g. XLM-R, but use a random initialization;
    \item The student is trained on 2M sentences of English monolingual data, and we also add 2M sentence of English-Spanish bitexts from CCMatrix to better align with the teacher's multilingual embedding space;
    \item Instead of one massively multilingual model, we train multiple students for a small subset of (similar) languages, or even a single language;
    \item Use of separate SPM vocabularies for teacher and student, better accommodating scripts and tokens in the student languages which were unseen by the teacher (cf. \autoref{sec:student_specific_SPM})
    \item Optimization of the cosine loss between the teacher and student embedding, since this is the relevant metric for bitext mining (cf. \autoref{fig:archi} above);
    \item Jointly train distillation alongside a MLM criterion to benefit additional learning from monolingual data in a foreign language (cf.  \autoref{fig:archi}, and \autoref{sec:joint_training}).
    \item Addition of curriculum learning in the form of \emph{progressive distillation}. In this strategy, instead of sending the entire sentence pairs all at once, we send incremental versions of the respective sentence pairs to both teacher and student, which we found to be helpful for some particularly challenging low-resource languages.
\end{itemize}

\noindent Our motivation of using a total of 4M English sentences is to \textit{``anchor''} the student encoder to the embedding space, and hopefully be able to learn new languages with limited amount of parallel texts.

In initial experiments, we used a 6-layer \mbox{BiLSTM} encoder architecture as in \citet{Artetxe:2019:tacl_massive_ml}, but we saw consistent improvements by switching to a 12-layer transformer.
We keep the same student architecture for all languages (L=12, H=1024, A=4, 250M params). When we minimize the cosine distance only, max-pooling of the transformer outputs to achieve the fixed-size sentence representations worked best, compared to using a special token like \texttt{[CLS]}. For curriculum learning using \emph{progressive distillation}, we incrementally send a percentage of subwords from each sentence pair (e.g. 10\%, 20\%, ..., 100\%). We experimented sending various incremental percentages of the sentence pairs to both teacher and student
(e.g. 20\%, 40\%), but found 10\% increments to perform best.
Teacher-student training was performed on 16 GPUs, ADAM optimizer, a learning rate of 0.0005 and with a batch size of 10,000.

\section{Training and evaluation resources}
\label{sec:ressources}

The sentence encoders are trained on publicly available bitexts, in particular from OPUS.\footnote{\url{https://opus.nlpl.eu/}}
Our monolingual data comes mostly from Common Crawl and other public sources like ParaCrawl\footnote{\url{https://paracrawl.eu}}, and some additional targeted
crawling for several low-resource languages. Preprocessing includes the following steps: sentence splitting, filtering of sentences in the wrong script or with more than 20\% of numbers or punctuation, LID and deduplication, as well as LM filtering on English.  
We have extended and improved fastText LID \cite{fasttextlid:2018:arxiv} to include additional languages
considered in this work. We trained this new LID model on publicly available monolingual data and evaluated it on human curated labeled test set of FLORES. 
The available monolingual and bitext resources are summarized in the result sections.

Creating high-quality development and test data for low-resource languages is
challenging. In this work, in order to evaluate our approach we make use of two publicly available corpora: Tatoeba and FLORES. The Tatoeba corpus\footnotemark[1] is a test set covering 112 languages \citep{Artetxe:2019:tacl_massive_ml}, and contains up to 1000 sentences for each language pair. Flores101 is a corpus made publicly available by the FLORES
project \cite{flores101:2021}. It covers 101 languages, and contains 1012 sentences for each language pair in the devtest set.\footnote{\url{https://github.com/facebookresearch/flores}}
We extended FLORES to \nbLangFlores{} African languages on which we report results in this paper, and will make the extended datasets freely available in the next months.

\begin{table*}[t!]
\centering
\begin{tabular}{ccccc|cccc}
\toprule
\multicolumn{1}{l}{{ }} & \multicolumn{1}{l}{{ }} & \multicolumn{3}{c}{{ \textbf{FLORES}}}                                                                                 & \multicolumn{4}{c}{{ \textbf{Tatoeba}}}                                                                                \\
\cmidrule{3-9}
{ \textbf{ISO}}         & { \textbf{Language}}    & \multicolumn{1}{c}{{ \textbf{LASER}}} & { \textbf{LASER3}} & { \textbf{LaBSE}} & \multicolumn{1}{c}{{ \textbf{LASER}}} & { \textbf{LASER3}} & { \textbf{LaBSE}}  & \textbf{\# Sents}\\
\midrule
{ amh}                  & { Amharic}              & {57.4}                                   & { 0.1}             & { 0}              & {51.2}                                   & { 10.7}           & { 5.4}           & 168                                   \\
{ bel}                  & { Belarusian}           & {40.4}                                   & { 0.3}             & { 0}              & {29.6}                                   & { 5.2}             & { 3.1}            & 1000                                  \\
{ gle}                  & { Irish}                & {92.5}                                   & { 0.8}             & { 0}              & {94.9}                                   & { 15.8}            & { 3.5}            & 1000                                  \\
{ hye}                  & { Armenian}             & {75.6}                                   & { 0.2}             & { 0}              & {59.8}                                   & { 8.0}             & { 3.8}           & 742                                   \\
{ kat}                  & { Georgian}             & {61.2}                                   & { 1.8}            & { 0}              & {60.3}                                   & { 21.2}           & { 3.6}           & 746                                   \\
{ kaz}                  & { Kazakh}               & \multicolumn{1}{c}{{63.3}}               & { 0.5}             & { 0.2}            & {79.3}                                   & { 16.7}           & { 8.7}            & 575                                   \\
{ khm}                  & { Khmer}        & \multicolumn{1}{c}{{64.3}}               & { 2.1}            & { 2.0}              & {74.0}                                   & { 43.6}           & { 15.0}          & 722                                   \\
{ swh}                  & { Swahili}              & {0.8}                                   & { 0.1}               & { 0}              & {36.7}                                   & { 16.9}           & { 9.0}           & 390                                   \\
{ tam}                  & { Tamil}                & {40.6}                                   & { 0.2}             & { 0}              & {23.1}                                   & { 37.8}           & { 6.5}           & 307                                   \\
{ tel}                  & { Telugu}               & {6.8}                                   & { 0.2}             & { 0}              & {16.2}                                   & { 17.1}           & { 1.7}           & 234                                   \\
{ urd}                  & { Urdu}                 & \multicolumn{1}{c}{{6.7}}               & { 0.2}             & { 0.1}            & {12.5}                                   & { 9.1}             & { 3.6}            & 1000                                  \\
{ uzb}                  & { Uzbek}                & \multicolumn{1}{c}{{79.9}}               & { 0.2}             & { 0.1}            & {77.3}                                   & { 18.2}            & { 10.5}          & 428                                   \\
\midrule
\multicolumn{1}{l}{{ }} & { \textbf{Average}}     & {49.1}                                   & { 0.6}            & { 0.2}           & {51.2}                                   & { 18.4}           & { 6.3}           & \multicolumn{1}{l}{}    \\ 
\bottomrule
\end{tabular}

\caption{Comparison of LASER, LASER3, and LaBSE on both FLORES and Tatoeba test sets (\xsim{} error rates). FLORES devtest has 1012 sentences and is N-way parallel.}
\label{tab:improving_laser}
\end{table*}

\subsection{Multilingual similarity search for mining}
\label{sec:xsim_tool}
In order to mine for bitexts, one approach is to compute the cosine similarity of sentence pairs so that parallel sentences can be obtained using nearest neighbor retrieval, and then subsequently filtered by setting a fixed threshold over the similarity score \citep{Schwenk:2018:acl_mine}.
However, it was shown that this approach suffers from scale inconsistency issues \citep{Guo:2018:arxiv_mine_bilingual}. 
To address this, \citet{Artetxe:2019:mine_acl} suggest a margin-based similarity method, in this work referred to as \xsim{}. It is defined as:
\begin{multline}
    \xsim(x,y) = \text{margin}( cos(x,y), \\ 
    \sum_{z \in NN_k(x)} \frac{cos(x, z)}{2k} + \sum_{z \in NN_k(y)} \frac{cos(y, z)}{2k} )
\end{multline}

\noindent where $x$ and $y$ are the source and target sentences,
and $NN_k(x)$ denotes the $k$ nearest neighbors of $x$
in the other language. There are three different margin functions: \emph{absolute} ($\text{margin}(a, b) = a$), \emph{ratio} ($\text{margin}(a, b) = \frac{a}{b}$), and \emph{distance} ($\text{margin}(a, b) = a - b$). As our end goal in this work is to produce encoders for the task of bitext mining, we adopt this approach, and evaluate all encoders using \xsim{} error rate with \emph{distance} margin.\footnote{All results are calculated into English (i.e. xxx $\rightarrow$ eng) and for LASER3 only, xxx is encoded by the student and eng is encoded by the teacher.}

\section{Experimental evaluation: multilingual similarity search}

\label{sec:eval_xsim}
In this section, we provide some evaluations of our proposed multilingual distillation approach, based on multilingual similarity search. We first show a comparison of our student encoders to the original LASER encoder, and also highlight student encoders trained with language-specific vocabularies, and the effect of jointly training our students using masked language modelling and curriculum learning.

\subsection{Improving LASER}

Given that LASER has been shown to perform well on many languages already, rather than focusing on marginal improvements for these languages, we instead selected several languages for which the original LASER encoder had an average accuracy of less than $90\%$ on the Tatoeba test set. However, as the Tatoeba test set is translated by volunteers, contains a majority of easy confusable short sentences, and for some languages has much less than 1000 sentences, we propose in this work to instead primarily rely on the FLORES dataset as the ground truth. This dataset is of a higher quality as a result of professional human annotation, and contains the same number of sentences across languages. Also, FLORES is N-way parallel and the results are comparable among languages. To illustrate this difference between datasets, we provide results in \autoref{tab:improving_laser} for the same languages across both test sets. 

In all instances on FLORES, we observe significant improvements upon the original LASER encoders using our proposed teacher-student approach, and also achieve competitive results to the much larger \emph{one-for-all} model LaBSE (average difference of 0.4\% \xsim{} error rate) 
We also notice that there is a considerable difference between both test sets. For example, on FLORES we report an \xsim{} of 0.1\% for Swahili, but an \xsim{} of 16.9\% of Tatoeba. To see if this phenomenon occurs with other representations, we also included LaBSE, for which we observed a similar effect. This stark difference further suggests that Tatoeba is a less reliable benchmark for evaluating sentence encoders. In particular, Tatoeba mainly contains very short sentences which can create a strong bias towards a particular model or training corpus. Given this observation, in the rest of this work we move away from Tatoeba and instead evaluate on FLORES.
We hope that other existing approaches and future work will also adopt evaluation on FLORES using a margin criterion. 

Although we also hoped to show a comparison to a similar distillation method by \citet{Reimers:2020:emnlp_destil}, their existing results were reported on Tatoeba (which as shown above is not very reliable to compare against), and results were not reported using the margin score (cf. \autoref{sec:xsim_tool}). We attempted to evaluate their reported models on FLORES using \emph{distance} margin, but their model is not available. We also attempted to reproduce the author's result by training new models using the provided code, but as we were not able to obtain the original training data used, we were unfortunately not able to reach a reasonably close outcome to make a fair comparison.
 
\begin{table}[t!]
    \begin{tabular}{lcrrr}
        \toprule
         \bf Training & \bf SPM & \bf \#train & \bf amh & \bf tir \\
         \midrule
         
         LASER2 & 50k joint & 220M & 34.88 & 92.89 \\
         Semitic & 50k joint & 9M & 0.49 & 1.68 \\
         Ge'ez & 8k specific & 0.7M & 0.1 & 0.89 \\
         \midrule
         LaBSE & 501k joint & $\approx$ 6000M & 0 & 13.74 \\
         \bottomrule
    \end{tabular}
    \caption{\xsim{} error rates on FLORES devtest for Amharic and Tigrinya and different training strategies (see text for details). The amount of training data excludes 4M sentences of English for our models.}
    \label{tab:geez}
\end{table}

\subsection{Language-specific encoders}
\label{sec:student_specific_SPM}

In our first experiments, we used the same preprocessing and SPM vocabulary for each student as the LASER2 SPM teacher. In particular, a 50k SPM vocabulary which was trained on all LASER2 languages. On one hand, using a massively multilingual SPM vocabulary is expected to improve the generalization among languages, since they may share several SPM tokens and it is the only possible solution when training a massively multilingual model which handles all languages.
On the other hand, low-resource languages may be badly modeled in a joint SPM vocabulary, i.e. mostly by very short SPM tokens, despite the use of up-sampling strategies.
Our approach to train multiple sentence encoders, each one specific to a small number of languages, opens the possibility to train and use specific SPM vocabularies for each subset of a small number of languages.
Table~\ref{tab:geez} summarizes the results for these different training strategies for some example languages: Amharic (amh) and Tigrinya (tir). Both are part of the family of Semitic languages, and use their own specific Ge'ez script. Other major languages from this family are Aramaic, Arabic and Hebrew, Maltese and Tigre, all using their own specific script.

Amharic was part of the \nbLangLaser{} languages LASER was trained on, but the \xsim{} error rate is rather high. LASER2 generalizes badly to Tigrinya. We first trained a specific encoder for three Semitic languages: Amharic, Tigrinya and Maltese. We only added Maltese, which uses a Latin script, in order to avoid a multitude of different scripts to be learnt by one encoder. This yields a significant improvement to 0.2 and 1.19\% respectively, highlighting the usefulness of teacher-student training and specific encoders for a small set of similar languages.
We then trained an encoder for Amharic and Tigrinya only, paired with English as in all our experiments, and a specific 8k SPM vocabulary to better support the Ge'ez script. This brought \xsim{} down to 0.1\% and 0.89\%, respectively although we use less training data.
Our best model is on par with LaBSE, which was trained on Amharic only, and significantly outperforms it for Tigrinya.

\begin{table}[!t]
\centering
\begin{tabular}{lr}
\toprule
\bf Approach & \xsim{} \\
\midrule
LASER & 70.65\\
LaBSE & 26.28\\
\midrule
LASER3 & 21.05 \\
+MLM & 12.65 \\
+MLM + Curriculum learning & 6.03\\
\bottomrule
\end{tabular}
\caption{Comparison of LASER and LaBSE to Wolof student models trained with and without MLM and curriculum (\xsim{} error rates).}
\label{tab:mlm_ablation}
\end{table}

\subsection{Joint training}
\label{sec:joint_training}

In order to highlight the effect of jointly training our students with masked language modelling and curriculum learning, we chose a very low-resource language with little bitexts available to use for distillation alone: Wolof.  As with previous students, we trained Wolof alongside closely related Senegambian languages: Fulah, Bassari, and Wamey, but the joint training strategies are only applied to Wolof. In total we used ~21k bitexts, and an additional 94k of monolingual data for Wolof. 
Results are shown in \autoref{tab:mlm_ablation}.

We observe a large reduction in \xsim{} when using joint training. For example, we see a 40\% relative reduction when adding the MLM criterion \mbox{(21.05 $\rightarrow$ 12.65)}, and a further decrease of \mbox{12.65 $\rightarrow$ 6.03} when also adding in curriculum learning. As we also observed a similar effect for other languages, the results above suggest that jointly training distillation alongside masked language modelling and curriculum learning is particularly beneficial for such low resource languages.

\section{Encoding and mining very low-resource languages}
\label{sec:eval_afric}

About 1.2 billion people are living in Africa, and with an estimated number of 1000--2000 languages, Africa is home to approximately one-third of the world's languages.
However, to the best of our knowledge, less than fifteen African languages are currently handled by public MT systems.
Most of the African languages are considered as very low-resource languages, i.e. less than 100 thousand sentence of bitexts are publicly available. Those resources are mainly religious texts, e.g. Bible translations,
which can lead to a domain mismatch when directly training NMT systems on this data.

In this section, we investigate the challenges to train sentence encoders for \nbLangAfric{} African languages, perform bitext mining, and train NMT models to translate between all these African languages and English and French, respectively.
The models and resources for \nbLangWMT{} languages are available in the framework of the WMT'22 workshop on large-scale translation of African languages.\footnote{\tiny \url{https://www.statmt.org/wmt22/large-scale-multilingual-translation-task.html}} More resources will be published within the next months.

\subsection{Choice of African languages}

We tried to cover as many African languages as possible. The main limitation was the availability of high-quality test sets to evaluate our models. In this work, we use FLORES \cite{flores101:2021}. It is available for the \nbLangWMT{} languages of the WMT'22 workshop on African languages: afr, amh, fuv, hau, ibo, kam, kin, lin, lug, luo, nso, nya, orm, sna, som, ssw, swh, tsn, tso, umb, wol, xho, yor and zul.
We extended FLORES to \nbLangFlores{} African languages on which we report results in this paper, and will make the extended datasets freely available in the next months.
Finally, we added \nbLangAddtl{} languages for which we have no FLORES test sets, namely Acholi, Luba, Luvale, Tiv, Venda and Zande, but sufficient resources to train sentence encoders and NMT systems.
Statistics for all \nbLangFlores{} languages are given in \autoref{tab:afric}.

\begin{table*}[p]
  
\renewcommand{\arraystretch}{0.8}
\begin{tabular}[t]{|cc|rr|rr|rrr|}
\toprule
\bf ISO & \bf Language & \bf Bitexts & \bf Mono & \MC{2}{c|}{\bf xsim [\%]} & \bf Mined & \MC{2}{c|}{\bf BLEU xxx/eng} \\
 & & \MC{1}{c}{\bf [k]} & \MC{1}{c|}{\bf [k]} & \bf LabSE & \bf LASER3 & \MC{1}{c}{\bf [k]}  & \bf public & \bf +mined \\
\midrule
afr & Afrikanns & 2061 & 0 & \bf 0.00 & 0.00 & 24240 & 50.72 & 55.15 \\
aka & Akan & 13 & 0 & 27.57 & 0.40 & 533 & 0.15 & 2.31 \\
amh & Amharic & 448 & 0 & \bf 0.00 & 0.10 & 9267 & 14.87 & 27.00 \\
bam & Bambara & 16 & 4 & 40.61 & 4.74 & 656 & 0.61 & 3.80 \\
bem & Bemba & 700 & 0 & 12.25 & 0.10 & 2166 & 15.38 & 17.71 \\
cjk & Chokwe & 40 & 16 & 34.39 & 16.40 & 839 & 0.00 & 2.02 \\
dik & Dinka & 25 & 21 & 37.94 & 21.84 & 571 & 0.00 & 2.70 \\
dyu & Dyula & 67 & 21 & 47.23 & 21.15 & 1177 & 0.48 & 1.30 \\
ewe & Ewe & 642 & 1 & 39.03 & 1.28 & 3057 & 11.30 & 11.33 \\
fon & Fon & 44 & 14 & 48.52 & 14.43 & 1009 & 1.15 & 2.67 \\
fuv & Fulfulde & 26 & 28 & 32.51 & 28.46 & 4509 & 0.00 & 6.52 \\
hau & Hausa & 416 & 0 & \bf 0.30 & 0.59 & 8454 & 19.22 & 29.67 \\
ibo & Igbo & 524 & 0 & \bf 0.00 & 0.20 & 5618 & 17.98 & 21.74 \\
kam & Kamba & 58 & 15 & 27.37 & 15.32 & 948 & 1.43 & 2.75 \\
kau\_Arab & Kanuri & 6 & 60 & 74.80 & 60.18 & 3866 & 0.00 & 1.11 \\
kau\_Latn & Kanuri & 11 & 4 & 37.65 & 4.64 & 307 & 0.00 & 2.58 \\
kik & Kikuyu & 119 & 1 & 27.27 & 1.28 & 1416 & 5.26 & 8.25 \\
kin & Kinyarwanda & 2012 & 0 & \bf 0.20 & 0.30 & 8385 & 17.76 & 20.70 \\
kmb & Kimbundu & 101 & 7 & 34.98 & 7.51 & 875 & 2.10 & 3.04 \\
kon & Kongo & 229 & 0 & 24.21 & 0.99 & 1497 & 7.83 & 9.09 \\
lin & Lingala & 1038 & 0 & 22.83 & 0.40 & 2632 & 16.40 & 16.94 \\
lua & Luba-Kasai & 325 & 1 & 24.90 & 1.98 & 1635 & 6.83 & 8.14 \\
lug & Luganda & 304 & 1 & 13.34 & 1.19 & 2901 & 9.07 & 12.55 \\
luo & Luo & 158 & 0 & 35.57 & 0.49 & 2244 & 6.60 & 11.50 \\
nso & Northen Sotho & 624 & 0 & 0.30 & 0.20 & 2526 & 23.06 & 27.62 \\
nus & Nuer & 21 & 7 & 50.40 & 7.21 & 785 & 0.00 & 3.28 \\
nya & Chewa; Nyanja & 867 & 0 & \bf 0.00 & 0.20 & 6301 & 17.94 & 22.55 \\
orm & Oromo & 177 & 0 & 45.75 & 0.49 & 1916 & 5.65 & 9.52 \\
run & Rundi & 665 & 0 & 0.10 & 0.49 & 3428 & 12.58 & 16.22 \\
sna & Shona & 826 & 0 & \bf 0.30 & 0.30 & 5959 & 19.57 & 22.90 \\
som & Somali & 179 & 0 & \bf 0.20 & 0.69 & 4935 & 5.13 & 21.30 \\
sot & Sotho & 1515 & 0 & \bf 0.00 & 0.10 & 6326 & 23.16 & 30.96 \\
ssw & Swati & 436 & 0 & 2.08 & 0.40 & 1407 & 6.88 & 15.14 \\
swh & Swahili & 1871 & 0 & \bf 0.00 & 0.10 & 14238 & 32.41 & 38.57 \\
tir & Tigrinya & 115 & 0 & 13.74 & 0.89 & 2380 & 3.60 & 12.04 \\
tsn & Tswana & 899 & 1 & 1.28 & 1.19 & 4298 & 20.09 & 20.63 \\
tso & Tsonga & 851 & 0 & 22.73 & 0.79 & 3294 & 22.36 & 23.65 \\
tum & Tumbuka & 585 & 1 & 5.43 & 1.68 & 2966 & 8.92 & 11.19 \\
twi & Twi & 630 & 0 & 24.60 & 0.69 & 2726 & 14.53 & 14.89 \\
umb & Umbundu & 233 & 15 & 36.96 & 15.61 & 1299 & 2.24 & 3.25 \\
wol & Wolof & 9 & 6 & \bf 26.28 & 6.03 & 808 & 0.00 & 3.09 \\
xho & Xhosa & 1176 & 0 & \bf 0.10 & 0.20 & 6315 & 26.80 & 31.92 \\
yor & Yoruba & 518 & 3 & \bf 0.69 & 3.66 & 5867 & 12.60 & 15.61 \\
zul & Zulu & 1758 & 0 & \bf 0.10 & 0.20 & 9167 & 29.45 & 33.86 \\
\bottomrule
\end{tabular}

  \caption{List of African languages, available resources and result summary. LaBSE's \xsim{} error rate in bold correspond to languages it was trained on. All results are on FLORES devtest.
  }
  \label{tab:afric}
\end{table*}

\subsection{Encoder training and evaluation}

We have explored several techniques to train sentence encoders on multiple languages, grouped into \textit{``families''} in different ways.
The largest family of African languages are by far Bantu languages. Other language families include Chadic, Cushtic, Kwa, Mande, Nilotic, Semitic and Senegambian.
We used only publicly available bitexts to train our sentence encoders.
We first trained one encoder on all African languages and then tried to improve them by using smaller language family specific models.
Unfortunately, several language families have a very small total amount of bitext training data, in particular Mande languages (83k) or Senegambian (36k). We were not able to train language specific encoders for these families which performed better than when trained together with all other African languages.
The following languages were trained separately:
\begin{itemize}
    \item Semitic: amh and tir
  \item Kwa languages: aka, ewe, fon and twi
  \item Wolof
\end{itemize}

\autoref{tab:afric} provides the \xsim{} scores for all languages for which we have FLORES devtest data. We always use the LASER2 teacher model for English and not the individual student models (which were also trained on English). This ensures that all students are mutually compatible and simplifies mining. We provide an analysis of \xsim{} between the languages and French in the Appendix.
For comparison, we also evaluated the publicly available LaBSE model\footnote{\url{https://github.com/bojone/labse}} on our test data. LaBSE was trained on a total of \nbLangLabse{} languages which includes \nbLangLabseAfric{} African languages (in bold in \autoref{tab:afric}). LaBSE performs very well on all of them, except Wolof which has \xsim{} of 26.3\%. Our encoder for Wolof achieves less than 10\% \xsim{} error. LabSE generalizes well to 5 other languages: Northern Sotho (nso), Rundi (run), Swati (ssw), Tswana (tsn) and Tumbuku (tum), LabSE's \xsim{} scores for the other languages are rather high.

Our LASER3 sentence encoders have \xsim{} error rates below 5\% for 33 languages.
The most difficult languages are: cjk, dik, dyu, fon, kam, kau, kmb, nus, umb and wol.
For most of them, we have a very limited amount of bitexts (less than 50k) and monolingual data (less than 1M).

\subsection{Bitext evaluation}

We now turn to using these encoders for bitext mining.
We follow exactly the same margin-based mining procedure as described in \citet{Artetxe:2019:mine_acl}, but use the union of forward and back-ward mining as introduced in \citet{schwenk:2021:acl_ccmatrix}.
Our main source of monolingual data was Common Crawl, complemented and targeted crawling (see \autoref{sec:ressources} for details on preprocessing).
The amount of unique sentences is given in \autoref{tab:afric} in column "Mono [k]". We mine against \monoENG{} unique sentences in English.

\paragraph{NMT training}
To evaluate the quality of the mined bitexts, we add the mined bitexts to the available public bitexts and train NMT systems, translating form foreign into English, and compare the BLEU scores with baseline models which were trained on freely available (human translated) bitexts only.
We train NMT systems to translate separately from each language into English. We hope that this gives us signals on the quality of the mined bitexts. For simplicity, we use the same architecture for all languages: a 6 layer transformer for the encoder and decoder, 8 attention heads, ffn=4096 and 512-dimensional embeddings. Models were trained for 100 epochs on 32 GPUs
The results are summarized in \autoref{tab:afric}, last three columns.

\paragraph{Analysis.}
We observe significant gains in the BLEU scores for several languages, e.g. amh, fuv, hau, lua, sot, ssw, swh, tir, som and xho, all improve by more than 5 points BLEU.
The most impressive result is obtained for Somali: training an NMT system on the available 179k bitexts yields 5.1 BLEU. This is then improved to 21.3 BLEU by adding 4.9M mined bitexts.
We also obtain a nice result on Fulfude: publicly available bitexts are extremely limited (26k) and we are able to reach 6.5 BLEU using mined bitexts, despite a sentence encoder with a high \xsim{} error rate of 27.4\%.
There are 13 languages with BLEU scores below 5\%: aka, bam, cjk, dik, dyu, fon, kam, kau\_Arab, kau\_Latn, kmb, nus, umb and wol. The sentence encoders for most of these languages need to be improved further, but the limiting factor is often the amount of available monolingual data - we simply have not enough data to mine in.
A typical example is Akan (aka): we have a very good sentence encoder - since it was trained jointly with the other Kwa languages, but only 163k sentences of monolingual data. There is not much mining can do here.

We would like to emphasize that these results should not be considered as the best possible MT performance which can be obtained with the available resources. We made no attempt to optimize the precision/recall trade-off of the mining individually for each language pair, i.e. the margin threshold, nor did we adapt the NMT architecture and parameters to the amount of bitexts.
We also expect that significant improvements in the BLEU scores can be obtained by training a multilingual NMT jointly on all languages.

\section{Conclusion}

Massively multilingual sentence representations are key to extend NLP approaches to more languages, and they are the underlying engine for distance-based bitext mining, which turned out to be crucial to scale NMT to more languages.
In this work, we attack the challenge to scale the LASER encoder beyond the 100 most frequent languages, and cover \nbLangAfric{} African languages. To the best of our knowledge, only \nbLangLabseAfric{} African languages are handled and evaluated by current multilingual models.

We achieve this by moving away from a \textit{one-for-all} approach to an improved teacher-student training of several encoders, each one trained on a small subset of languages. This enabled us to better adapt the encoders to language specificities, e.g. a particular writing script, while maintaining mutual compatibility. 
Our new models significantly outperform the original LASER model on the FLORES test set, and we are on par or better than all other publicly available multilingual sentence encoders, namely LaBSE.
We were also able to integrate monolingual data by jointly minimizing a cosine and MLM loss. We showcase the potential of this technique for the Wolof language, reducing the \xsim{} error rate from 21.05\% down to 6.03\%.

We performed bitext mining for \nbLangFlores{} African languages and trained an NMT model which can translate all these languages from and into English and French, respectively.
The encoders and bitexts for \nbLangWMT{} languages are available in the framework of the EMNLP'22 workshop on Large-Scale Machine Translation Evaluation for African Languages.

\section{Acknowledgements}

For their helpful contributions to this work, we would like to thank: Bapi Akula, Pierre Andrews, Angela Fan, Cynthia Gao, Kenneth Heafield, Philipp Koehn, Janice Lam, Alex Mourachko, Christophe Rogers and Guillaume Wenzek.

\bibliographystyle{acl_natbib}
\bibliography{laser3_arxiv22}

\appendix

\begin{sidewaystable*}[p!]
  \centering
  \renewcommand{\arraystretch}{0.8}
\footnotesize
\begin{tabular}[t]{@{}c@{\,}|*{33}{p{4pt}}}
\toprule
& eng & fra & afr & aka & amh & bem & ewe & hau & ibo & kik & kin & kon & lin & lua & lug & luo & nso & nya & orm & run & sna & som & sot & ssw & swh & tir & tsn & tso & tum & twi & xho & zul \\
\midrule
eng  &   & 0.0  & 0.0  & 1.0  & 0.1  & 0.1  & 2.8  & 0.8  & 0.5  & 0.7  & 0.3  & 1.5  & 0.2  & 1.8  & 2.6  & 0.8  & 0.1  & 0.2  & 1.3  & 0.4  & 0.2  & 2.0  & 0.1  & 0.4  & 0.9  & 0.6  & 1.4  & 0.7  & 1.7  & 0.4  & 0.1  & 0.1 \\
fra  & 0.0  &   & 0.0  & 1.6  & 0.1  & 1.0  & 2.9  & 1.1  & 0.4  & 0.9  & 0.4  & 1.6  & 0.2  & 2.6  & 3.5  & 0.9  & 0.4  & 0.6  & 2.5  & 0.6  & 0.2  & 2.2  & 0.2  & 0.4  & 1.7  & 1.0  & 1.3  & 0.6  & 2.5  & 0.8  & 0.1  & 0.1 \\
afr  & 0.0  & 0.2  &   & 5.7  & 1.5  & 3.2  & 9.0  & 3.5  & 3.4  & 4.0  & 2.5  & 6.9  & 3.1  & 7.6  & 8.1  & 3.5  & 2.3  & 3.3  & 10  & 4.5  & 2.5  & 6.5  & 1.5  & 3.4  & 4.5  & 7.7  & 4.5  & 3.1  & 7.4  & 5.0  & 1.3  & 1.2 \\
aka  & 0.4  & 1.1  & 3.6  &   & 2.9  & 2.4  & 7.7  & 4.6  & 3.4  & 4.0  & 2.9  & 5.2  & 2.5  & 9.3  & 8.7  & 5.0  & 1.5  & 3.7  & 8.7  & 3.7  & 3.0  & 7.0  & 1.8  & 2.9  & 3.3  & 8.0  & 3.3  & 2.3  & 5.8  & 2.0  & 2.2  & 1.3 \\
amh  & 0.1  & 0.2  & 1.1  & 4.8  &   & 1.9  & 5.8  & 3.4  & 2.4  & 2.2  & 1.3  & 4.2  & 2.1  & 7.0  & 6.8  & 2.5  & 1.1  & 2.0  & 5.0  & 2.7  & 1.0  & 4.5  & 0.4  & 1.6  & 4.2  & 2.7  & 3.0  & 2.8  & 5.2  & 2.1  & 0.5  & 0.7 \\
bem  & 0.1  & 0.6  & 1.8  & 2.7  & 1.1  &   & 6.2  & 2.7  & 2.1  & 2.2  & 1.1  & 3.3  & 1.6  & 4.3  & 5.8  & 1.9  & 0.4  & 1.1  & 5.3  & 1.4  & 1.2  & 4.0  & 0.6  & 1.2  & 1.7  & 5.1  & 1.8  & 2.1  & 4.2  & 2.7  & 0.9  & 0.3 \\
ewe  & 1.3  & 1.7  & 5.1  & 7.0  & 4.7  & 4.5  &   & 6.0  & 6.8  & 5.6  & 4.5  & 7.7  & 4.6  & 12  & 11  & 6.1  & 4.3  & 4.5  & 13  & 6.3  & 4.7  & 10.0  & 3.0  & 4.3  & 11  & 11  & 5.0  & 4.2  & 7.6  & 5.6  & 3.0  & 3.8 \\
hau  & 0.6  & 0.7  & 1.6  & 3.6  & 2.0  & 1.8  & 6.8  &   & 2.1  & 2.2  & 2.1  & 4.5  & 1.8  & 6.2  & 5.3  & 3.1  & 1.4  & 1.3  & 5.3  & 2.4  & 1.7  & 3.9  & 0.9  & 1.7  & 4.0  & 5.8  & 2.6  & 2.4  & 4.8  & 2.7  & 1.2  & 1.0 \\
ibo  & 0.2  & 0.4  & 1.8  & 3.6  & 1.6  & 2.2  & 6.6  & 2.3  &   & 2.8  & 2.2  & 3.9  & 2.1  & 6.7  & 4.9  & 3.7  & 1.2  & 2.4  & 5.7  & 2.7  & 2.2  & 4.5  & 1.5  & 2.1  & 2.9  & 5.7  & 2.7  & 2.3  & 5.5  & 2.4  & 1.4  & 1.0 \\
kik  & 1.3  & 1.4  & 2.6  & 5.7  & 2.5  & 2.5  & 7.6  & 3.1  & 2.9  &   & 2.8  & 5.3  & 2.7  & 6.0  & 6.6  & 3.2  & 1.8  & 3.0  & 7.9  & 4.0  & 2.7  & 5.5  & 1.7  & 2.5  & 2.5  & 6.8  & 3.1  & 2.5  & 4.8  & 4.0  & 1.5  & 1.3 \\
kin  & 0.3  & 0.5  & 1.8  & 2.8  & 1.0  & 1.6  & 6.4  & 2.6  & 1.7  & 2.6  &   & 2.7  & 1.1  & 5.3  & 4.9  & 2.3  & 0.7  & 1.4  & 4.5  & 1.8  & 1.5  & 3.6  & 0.9  & 1.2  & 1.9  & 3.2  & 2.6  & 1.3  & 4.2  & 2.0  & 0.6  & 0.6 \\
kon  & 1.0  & 0.5  & 4.3  & 5.1  & 3.4  & 3.7  & 8.7  & 5.6  & 5.1  & 4.8  & 4.2  &   & 3.1  & 9.2  & 8.2  & 4.9  & 2.9  & 4.0  & 11  & 5.1  & 2.9  & 6.2  & 2.2  & 3.0  & 3.5  & 9.3  & 4.5  & 3.6  & 6.5  & 5.3  & 1.9  & 2.5 \\
lin  & 0.4  & 0.4  & 1.6  & 2.6  & 1.3  & 1.3  & 4.9  & 2.6  & 2.3  & 3.0  & 1.2  & 2.4  &   & 5.5  & 5.3  & 2.0  & 1.0  & 1.3  & 5.1  & 2.0  & 1.0  & 4.3  & 0.6  & 1.3  & 2.2  & 4.0  & 2.1  & 1.7  & 3.8  & 1.7  & 0.4  & 0.6 \\
lua  & 2.0  & 2.6  & 5.5  & 8.7  & 6.8  & 4.3  & 13  & 6.7  & 7.1  & 6.1  & 5.9  & 8.0  & 5.6  &   & 10  & 6.9  & 4.2  & 5.8  & 14  & 8.7  & 5.6  & 8.5  & 4.7  & 5.6  & 6.5  & 13  & 5.5  & 5.8  & 9.5  & 7.5  & 4.0  & 3.9 \\
lug  & 1.2  & 1.9  & 4.0  & 8.9  & 4.2  & 4.2  & 10  & 6.8  & 5.4  & 5.0  & 4.7  & 7.4  & 4.8  & 11  &   & 5.5  & 4.0  & 4.2  & 13  & 6.8  & 3.9  & 8.3  & 2.6  & 4.0  & 5.7  & 13  & 4.9  & 6.1  & 9.9  & 6.6  & 3.0  & 3.6 \\
luo  & 0.5  & 0.9  & 3.3  & 7.1  & 3.5  & 3.4  & 10  & 5.9  & 5.0  & 3.8  & 3.9  & 6.3  & 4.2  & 9.5  & 9.9  &   & 3.7  & 4.5  & 11  & 5.6  & 4.5  & 7.3  & 2.1  & 3.4  & 4.3  & 12  & 4.2  & 3.7  & 7.8  & 5.5  & 1.9  & 2.8 \\
nso  & 0.2  & 0.5  & 0.6  & 2.2  & 0.7  & 1.0  & 3.9  & 1.8  & 1.7  & 1.4  & 0.7  & 2.4  & 1.0  & 4.3  & 3.5  & 1.7  &   & 0.6  & 4.0  & 1.5  & 0.7  & 3.2  & 0.4  & 0.5  & 0.9  & 2.3  & 1.8  & 1.3  & 3.4  & 1.3  & 0.4  & 0.3 \\
nya  & 0.2  & 0.5  & 1.7  & 3.3  & 1.0  & 1.2  & 6.0  & 2.4  & 3.1  & 2.7  & 1.3  & 2.9  & 1.4  & 5.5  & 5.7  & 2.5  & 1.4  &   & 5.3  & 3.0  & 0.8  & 4.0  & 1.0  & 1.1  & 2.4  & 4.3  & 2.4  & 1.7  & 3.0  & 2.7  & 0.4  & 0.5 \\
orm  & 0.5  & 1.5  & 4.9  & 8.6  & 2.8  & 4.0  & 13  & 7.3  & 5.3  & 5.6  & 6.1  & 8.6  & 5.6  & 14  & 14  & 8.3  & 4.7  & 5.4  &   & 8.8  & 6.8  & 9.2  & 4.0  & 6.1  & 4.5  & 10  & 5.9  & 5.7  & 9.9  & 6.4  & 4.0  & 3.3 \\
run  & 0.5  & 0.5  & 2.3  & 3.6  & 1.1  & 1.2  & 6.5  & 2.5  & 2.3  & 2.5  & 0.9  & 3.3  & 1.8  & 7.6  & 5.7  & 2.8  & 1.0  & 2.1  & 6.6  &   & 1.1  & 4.6  & 0.4  & 1.3  & 2.8  & 4.7  & 2.3  & 1.8  & 4.2  & 2.3  & 1.0  & 0.5 \\
sna  & 0.3  & 0.9  & 2.5  & 5.2  & 1.7  & 2.1  & 6.5  & 3.3  & 2.6  & 2.8  & 2.0  & 3.4  & 1.9  & 6.5  & 6.6  & 3.1  & 1.1  & 1.3  & 6.6  & 2.3  &   & 4.2  & 1.3  & 1.1  & 3.0  & 5.4  & 2.9  & 2.1  & 4.5  & 2.2  & 1.1  & 0.6 \\
som  & 0.7  & 0.5  & 3.1  & 6.5  & 1.4  & 3.2  & 9.6  & 4.0  & 3.5  & 4.2  & 3.7  & 4.8  & 3.6  & 9.5  & 8.1  & 4.5  & 2.4  & 3.7  & 6.7  & 5.1  & 3.0  &   & 2.0  & 2.8  & 2.3  & 7.4  & 4.0  & 3.8  & 7.6  & 5.0  & 2.4  & 1.2 \\
sot  & 0.1  & 0.0  & 1.1  & 2.6  & 0.7  & 0.7  & 4.6  & 1.8  & 0.8  & 1.6  & 0.8  & 2.0  & 0.7  & 4.9  & 4.0  & 1.0  & 0.4  & 0.8  & 3.6  & 1.4  & 0.8  & 3.1  &   & 0.7  & 1.6  & 3.1  & 1.5  & 1.4  & 3.0  & 1.6  & 0.2  & 0.1 \\
ssw  & 0.4  & 0.5  & 1.9  & 3.3  & 1.3  & 1.6  & 5.2  & 2.6  & 2.4  & 2.6  & 1.2  & 2.6  & 2.0  & 5.5  & 4.9  & 2.3  & 1.0  & 1.3  & 5.9  & 2.3  & 1.1  & 4.2  & 1.2  &   & 2.4  & 4.3  & 2.1  & 1.9  & 4.0  & 2.6  & 0.7  & 0.7 \\
swh  & 0.1  & 0.4  & 1.1  & 4.2  & 0.6  & 1.1  & 7.4  & 2.3  & 1.2  & 1.7  & 1.1  & 3.4  & 2.3  & 5.9  & 4.7  & 2.7  & 1.0  & 2.1  & 5.6  & 2.3  & 1.5  & 2.7  & 0.5  & 1.2  &   & 4.3  & 2.1  & 2.0  & 4.3  & 2.6  & 0.3  & 0.3 \\
tir  & 0.9  & 1.1  & 3.1  & 6.4  & 1.4  & 3.4  & 8.3  & 4.6  & 4.2  & 3.9  & 2.1  & 5.8  & 3.2  & 9.4  & 10  & 4.5  & 2.5  & 3.5  & 7.6  & 4.5  & 2.2  & 8.2  & 1.3  & 3.2  & 8.2  &   & 4.0  & 3.6  & 8.4  & 5.2  & 1.5  & 1.6 \\
tsn  & 1.2  & 1.3  & 2.6  & 4.5  & 2.1  & 2.3  & 6.3  & 4.0  & 2.8  & 3.0  & 2.9  & 3.8  & 2.4  & 6.0  & 6.0  & 3.5  & 1.9  & 2.3  & 6.3  & 3.1  & 2.4  & 4.5  & 1.9  & 2.3  & 2.5  & 4.5  &   & 2.3  & 4.5  & 3.6  & 1.9  & 1.5 \\
tso  & 0.8  & 0.8  & 1.4  & 2.2  & 1.5  & 1.6  & 4.5  & 3.0  & 1.9  & 2.0  & 1.4  & 3.2  & 1.4  & 5.7  & 6.1  & 3.2  & 0.8  & 1.3  & 4.3  & 2.3  & 1.3  & 4.2  & 0.9  & 1.3  & 2.3  & 3.5  & 2.2  &   & 3.2  & 1.7  & 0.8  & 0.7 \\
tum  & 1.7  & 1.6  & 4.0  & 6.5  & 3.7  & 3.3  & 7.8  & 5.2  & 5.0  & 4.6  & 3.6  & 6.0  & 4.2  & 9.4  & 9.7  & 5.4  & 3.1  & 3.3  & 9.8  & 4.5  & 3.3  & 7.6  & 2.9  & 3.1  & 5.3  & 9.1  & 4.5  & 3.7  &   & 4.2  & 2.6  & 2.1 \\
twi  & 0.7  & 1.0  & 2.9  & 2.5  & 2.1  & 2.8  & 7.1  & 4.2  & 3.0  & 3.6  & 2.7  & 5.7  & 2.8  & 6.5  & 7.0  & 3.5  & 1.2  & 2.9  & 8.1  & 4.0  & 2.3  & 6.5  & 1.7  & 2.8  & 3.8  & 7.8  & 3.2  & 2.4  & 5.5  &   & 1.7  & 1.7 \\
xho  & 0.2  & 0.2  & 1.2  & 3.3  & 0.4  & 1.1  & 5.0  & 2.5  & 1.6  & 1.5  & 1.2  & 2.2  & 1.1  & 4.3  & 4.8  & 1.7  & 0.6  & 1.3  & 4.9  & 2.1  & 0.9  & 3.7  & 0.6  & 0.7  & 1.8  & 3.6  & 2.3  & 1.5  & 3.5  & 1.5  &   & 0.4 \\
zul  & 0.2  & 0.1  & 0.4  & 1.7  & 0.5  & 0.8  & 3.7  & 1.3  & 0.9  & 0.9  & 0.8  & 1.9  & 0.6  & 3.7  & 3.5  & 1.3  & 0.5  & 0.5  & 2.8  & 0.7  & 0.4  & 2.8  & 0.4  & 0.6  & 1.3  & 1.8  & 1.5  & 0.5  & 2.1  & 1.3  & 0.2  &  \\
\bottomrule
\end{tabular}
  \caption{\xsim{} matrix for subset of African languages, and English and French. All results are on FLORES devtest.}
  \label{tab:multi}
\end{sidewaystable*}

\newpage

\section{Analysis of zero-short performance of multiple student encoders}
\label{sec:xsim_multi}

We trained our students to minimize the \xsim{} score of each language with respect to the English LASER2 teacher. In order to best consider specificities of languages, several independent student models were trained:
\begin{itemize}
    \item Semitic languages: amh and tir
    \item Kwa languages: aka, ewe, fon, and twi
    \item Senegambian languages: Wolof
    \item remaining 43 languages
\end{itemize}
This means that for instance the students of the Semitic family have never seen any data of all the other languages. The only link is the common teacher.
\autoref{tab:multi} gives the \xsim{} scores for all possible pairs. 
To limit the size of the table, we consider only the 30 best performing languages i.e., those with the smallest \xsim{} scores. Please note that the table is not symmetric (e.g. \mbox{eng $\rightarrow$ kon $=1.5$}, while \mbox{kon $\rightarrow$ eng $= 1.0$}).

We observe that the \xsim{} scores amongst the African language pairs are higher than with English, but they stay rather low for most of the pairs (below 5\%).
As an example, let us consider the two student models for Semitic and Kwa languages. Both were trained on few languages with a small amount of bitexts. Still, we achieve reasonable \xsim{} scores among them: 
\mbox{aka $\rightarrow$ amh $=2.9$},
\mbox{amh $\rightarrow$ tir $=1.1$},
or \mbox{tir $\rightarrow$ twi $=5.2$}.

\paragraph{Zero-shot performance with French}
Finally, we also added the \xsim{} scores of all languages with respect to French, encoded by the LASER2 teacher. Please note that none of the student models were trained to minimize the cosine distance to French embeddings.
Nevertheless, we observe very low \xsim{} scores, and close to those with English. There are even some pairs for which the \xsim{} error rates to French are smaller than to English, namely ibo/fra, som/fra and zul/fra.
This means that we can use our student encoders to mine against all languages supported by the LASER2 encoder.

\end{document}